\documentclass[aps,prb,twocolumn,superscriptaddress]{revtex4}
\usepackage{graphicx}
\usepackage{amssymb}
\usepackage{amsmath}
\usepackage{graphicx}
\usepackage{epsfig}
\usepackage{color}
\usepackage{ulem}
\usepackage{hyperref}
\usepackage{appendix}

\DeclareMathOperator*{\argmin}{\arg\min}
\DeclareMathOperator*{\argmax}{\arg\max}
\newcommand{\norm}[1]{\left\| #1\right\|}

\begin{document}
\title{Learned Optimizers for Analytic Continuation}
\author{Dongchen Huang}
\affiliation{Beijing National Laboratory for Condensed Matter Physics and Institute of
Physics, Chinese Academy of Sciences, Beijing 100190, China}
\affiliation{School of Physical Sciences, University of Chinese Academy of Sciences, Beijing 100049, China}
\author{Yi-feng Yang}
\email[]{yifeng@iphy.ac.cn}
\affiliation{Beijing National Laboratory for Condensed Matter Physics and Institute of
Physics, Chinese Academy of Sciences, Beijing 100190, China}
\affiliation{School of Physical Sciences, University of Chinese Academy of Sciences, Beijing 100049, China}
\affiliation{Songshan Lake Materials Laboratory, Dongguan, Guangdong 523808, China}
\date{\today}
\begin{abstract}
Traditional maximum entropy and sparsity-based algorithms for analytic continuation often suffer from the ill-posed kernel matrix or demand tremendous computation time for parameter tuning. Here we propose a neural network method by convex optimization and replace the ill-posed inverse problem by a sequence of well-conditioned surrogate problems. After training, the learned optimizers are able to give a solution of high quality with low time cost and achieve higher parameter efficiency than heuristic fully-connected networks. The output can also be used as a neural default model to improve the maximum entropy for better performance. Our methods may be easily extended to other high-dimensional inverse problems via large-scale pretraining.
\end{abstract}
\maketitle

\section{Introduction}
Inverse problems appear in many perspectives of physics and machine learning, such as learning Hamiltonian in the classical \cite{Chow1968,Hinton1986,Bresler2015} or quantum sense \cite{Anshu2020,Bairey2019,Amin2018} and recovering sparse signal from noise measurements \cite{Donoho2006,Candes2005}. In quantum many-body problems, correlation functions are often computed in imaginary time \cite{Gull2011,Gubernatis2016,Wei2017,Hu2019,Hu2020} so that an analytic continuation has to be implemented to obtain the spectral function in real frequency in order to extract meaningful information. The analytic continuation is nothing but a linear inverse problem, which is, however, highly ill-posed and may have infinite unphysical solutions.

Many algorithms \cite{Gianluca2017} have been proposed to attack this problem, including the pad\'e approximation \cite{Han2017}, stochastic methods \cite{Sandvik1998,Sandvik2016,Ghanem2020a,Ghanem2020b},  maximum entropy methods \cite{JARRELL1996,Sim2018,Kraberger2017,Silver1990,Gunnarsson2010b,Reymbaut2015,Rumetshofer2019}, and the Nevanlinna method \cite{Fei2021}. Classical methods such as the pad\'e approximation \cite{Deisz1996} and the singular value decomposition (SVD) \cite{Gunnarsson2010} have been applied to the Hubbard model. But none of them holds for all situations and a case-by-case tuning is often needed. For instance, maximum entropy methods demand a highly empirical selection of prior distributions of the spectral function.

From the view of representation learning, high dimensional data of real world always have certain low-dimensional structures. For a high dimensional vector, the simplest low-dimensional structure is sparsity, which means that the vector may have many zero entries. The spectral function may therefore also have a sparse structure if it is properly discretized. This has motivated a transformation of the analytic continuation problem to a basis pursuit (BP) problem and inspired a line of work focusing on sparsity of the spectral function \cite{Ohzeki2018,YOSHIMI2019,Otsuki2020,Otsuki2017,Shinaoka2017}. Unfortunately, the power of sparsity-based methods is greatly limited by the ill-poseness of the Fermi kernel matrix.

In this work, we develop a neural network architecture for analytic continuation by further transforming the highly \textit{ill-posed} BP problem into a sequence of \textit{well-conditioned} surrogate problems. Rather than solving the original ill-posed problem by an optimizer with predefined weights given directly by the Fermi kernel, we introduce a \textit{learned optimizer} whose neural network structure can be derived from convex optimization of the well-conditioned problem sequence with adaptive weights.

This neural network avoids empirical design and shows higher parameter efficiency compared with heuristic fully-connected neural networks (FCNs) used in other works \cite{Fournier2020,Yoon2018}, namely, it needs fewer parameters (weights) to achieve the same accuracy as FCNs (see Appendix A for architecture and training details). It can also give a high quality approximate solution of the linear inverse problem with much less time cost than traditional maximum entropy methods (see Appendix B). Moreover, the two approaches can complement each other by taking advantage of neural network's strengths and treating its output as prior distributions of the maximum entropy, thus yielding an improved solution with better precision.

This paper is organized as follows. In section II, we first introduce the analytic continuation and sparsity-based methods under the framework of Bayesian inference and then propose the learned optimizers inspired by the connection between the fixed-point problem and the neutral networks. In section III, we give some details on the application, performance, and robustness of our methods, and propose a neural default model to improve the traditional maximum entropy methods. Section IV is a brief conclusion.

\section{Method}

\subsection{Analytic continuation and maximum entropy} 
We are dealing with the inverse problem to obtain the spectral function $A(\omega)$ in real frequency from the Green's function $G(\tau)$, 
\begin{equation}
    G(\tau) = \int_{-\infty}^{\infty} K(\tau,\omega)A(\omega) d\omega,
    \label{Eq:Analytical-Problem}
\end{equation}
where $\tau$ is the imaginary time. For a fermionic Green's function, the kernel $K(\tau,\omega)$ takes the form,
\begin{equation}
    K(\tau,\omega) = \frac{e^{-\tau \omega}}{1+e^{-\beta \omega}},
\end{equation} 
where $\beta$ is the inverse temperature. For analytic continuation, we first discretize Eq.~\eqref{Eq:Analytical-Problem} and get the linear inverse problem (in matrix form):
\begin{equation}
    g(\tau_i) = K(\tau_i,\omega_j)a(\omega_j), 
    \label{Eq:Linear-AC}
\end{equation}
where $i=1,\dots,N_\tau$ and $j=1,\dots,N_\omega$ mark the discrete points in imaginary time and real frequency, respectively, $g$ is the vectorized Green's function, $a$ is the vectorized spectral function, and $K(.,.)$ is a matrix of the Fermi kernel. 

The above problem can be solved using the Bayesian inference. The posterior distribution of the spectral function $a$ satisfies the Bayes' theorem, 
\begin{equation}
    P(a|g) = \frac{P(g|a)P(a)}{P(g)},
    \label{Eq:Posterior-A}
\end{equation}
from which a solution $a^*$ can be derived by maximum likelihood, 
\begin{equation}
a^* = \argmax_a P(a|g).
\label{Eq:likelihood}
\end{equation}
The maximum entropy and sparsity-based methods are just two special forms of its implementation with different choices of $P(g|a)$ and $P(a)$. The traditional maximum entropy methods \cite{JARRELL1996} choose $P(g|a) \propto e^{-\chi^2/2}$ and $P(a) \propto e^{-\alpha S}$, where $\chi^2 = (g-Ka)^T\Sigma^{-1}(g-Ka)$ denotes the reconstruction error, $\Sigma$ is the empirical covariance matrix, $K$ is the kernel matrix, and $S=\sum_i \Delta \omega_i  a(\omega_i)\log \frac{a(\omega_i)}{d(\omega_i)}$ is the Kullback-Leibler (KL) divergence between the spectral function and a prior default model $d(.)$ which is typically chosen to be the uniform or Gaussian distribution. Putting these back into Eqs.~\eqref{Eq:Posterior-A} and \eqref{Eq:likelihood} and considering that $P(g)$ is independent of $a$, we have immediately the maximum entropy formalism,
\begin{equation}
  a^* =\argmin_a \frac{\chi^2}{2} + \alpha S.
  \label{Eq:MaxEnt}
\end{equation}
Thus, the maximum entropy methods favor a solution of least deviation from the default model. The hyperparameter $\alpha$ can be adjusted in different ways \cite{Bryan1990,Bergeron2016}.

\subsection{Sparsity-based methods}
By contrast, the sparsity-based methods look for a vector solution with a maximal number of zero entries \cite{Otsuki2017}. We are then dealing with an optimization problem: $\min \|a\|_0$ s.t. $\|g - Ka\|_2 \leq \epsilon$, where $\|.\|_0$ is the $\ell_0$ norm that counts the number of non-zero entries in a vector, $\|.\|_2$ is the $\ell_2$ norm, and $\epsilon$ denotes an error tolerance. However, minimizing the $\ell_0$ norm is NP-hard \cite{Natarajan1995}. Fortunately, the $\ell_1$ norm provides a good replacement which is the largest convex function to approximate $\ell_0$ norm. We have then a surrogate optimization problem,
\begin{equation}
\min \|a\|_1\ \  s.t.\ \  \|g - Ka\|_2 \leq \epsilon, 
\label{Eq:L1}
\end{equation}
where $\norm{.}_1$ is the $\ell_1$ norm defined as the summation of the absolute values of all elements in the vector.

The above equations may be put in the same probabilistic framework as the maximum entropy, if we assume a Gaussian distribution with unit variance such that $P(g|a)\propto e^{-\frac{1}{2}\|g-Ka\|_2^2}$ and choose the Laplacian prior distribution, $P(a)\propto e^{-\lambda \norm{a}_1}$, where $\lambda$ is a positive hyperparameter. There may also be other choices for $P(a)$ promoting sparsity different from the current sparse regularization, namely, the $\ell_1$ term. Following the same derivation for Eq. \eqref{Eq:MaxEnt}, we arrive at a BP problem \cite{Tibshirani1996}, 
\begin{equation}
    a^* = \argmin_a \frac{1}{2}\|g-Ka\|_2^2 + \lambda \|a\|_1.
    \label{Eq:BP}
\end{equation}
Clearly, Eq. \eqref{Eq:L1} and Eq. \eqref{Eq:BP} are equivalent and $\lambda$ may be viewed as a Lagrangian multiplier for solving Eq. \eqref{Eq:L1}.

The sparsity assumption and such kind of problems have achieved huge amounts of successes in machine learning and signal processing. The BP problem is well-posed and guaranteed to recovery the exact spectral function as long as $a$ is sparse enough and the kernel matrix $K$ satisfies some fine properties. One of the most popular and simplest measure of the “fineness” of a matrix is mutual coherence \cite{Donoho03, Gribonval03}. We will discuss it later for the Fermi kernel matrix.
 
Since Eq.~\eqref{Eq:BP} is convex, many popular methods can be applied and all of them can converge to the global minimum. Its solution $a^*$ must satisfy the optimality condition $0 \in K^T(Ka^*-g) + \lambda\partial \norm{a^*}_1$, where the superscript $T$ denotes matrix transpose and $\partial (.)$ denotes the subdifferential \footnote{The subdifferential of a convex function $f(x)$ at any $x$ is defined as the collection of all subgradient $v$ satisfying $f(y)\ge f(x)+\langle v, y-x\rangle$ for all $y$. The subdifferential is an extension of the usual differential and may be applied even when the function is not smooth such as the $l_1$ norm. For example, $\partial \norm{x}_1=1$ for $x>0$, -1 for $x<0$, and $[-1,1]$ at $x=0$.}. Thus, for any $\tau>0$, we have 
\begin{equation}
  a^* - \tau K^T(Ka^*-g) \in a^* + \tau\lambda \partial \norm{a^*}_1.
  \label{Eq:subgradient}
\end{equation} 
On the other hand, for any convex function $F:\mathbb{R}^N \rightarrow (-\infty,\infty]$ and its induced proximal mapping\footnote{Geometrically, the proximal mapping can be viewed as a generalization of the projection. For example, if we choose $F(x) = \chi_C(x)$ to be the characteristic function taking value $0$ if $x\in C$ and $\infty$ if $x\notin C$ where $C$ is some constrained set, the proximal mapping becomes a projection $P_{\chi_C}(z) = \argmin_{x\in C}\norm{x-z}_2^2$ of $z$ into the set $C$. We can generalize the projection by replacing the characteristic function $\chi_C$ to a more general convex function.} $P_F(z) = \argmin_x F(x) + \frac{1}{2}\norm{x-z}_2^2$, we have
\begin{equation}
  z\in x+ \partial F(x).
  \label{Eq:subgradient-proximal mapping}
\end{equation}  
Combining Eqs. \eqref{Eq:subgradient} and \eqref{Eq:subgradient-proximal mapping} gives the correspondence: $x \rightarrow a^*$, $z\rightarrow   a^* - \tau K^T(Ka^*-g)$, and $F(x)\rightarrow \tau\lambda\norm{x}_1$. The identity $x=P_F(z)$ immediately implies the fixed-point equation \cite{Boyd2004},
\begin{equation}
  a^* = S_{\tau \lambda} (a^*-\tau K^T(Ka^*-g)),
        \label{Eq:fixed-point}
\end{equation}
where $S_{\tau \lambda}(.)$ is the soft-thresholding operator given by the proximal mapping $P_F(z)$ of the function $F(x)=\tau\lambda\norm{x}_1$. By definition, we have \footnote{For $x = \argmin_x \frac{1}{2}\norm{x-z}_2^2 + \tau\lambda \norm{x}_1$, the optimality condition gives $0\in (x-z)+ \tau \lambda \partial \norm{x}_1$. Thus, we have $0= x-z+\tau \lambda \Rightarrow x = z-\tau\lambda $ if $x>0$ or $z>\tau\lambda$ and $0= x-z-\tau \lambda \Rightarrow x = z+\tau\lambda $ if $x<0$ or $z<\tau\lambda$ . For $x=0$, because the subgradient of $\norm{x}_1$ at zero is the set $[-1,1]$, we have $x=0$ if $z\in \tau\lambda [-1,1]$ or equivalently $|z|\leq \tau\lambda$. Combining the above three situations gives $x=S_{\tau\lambda}(z)$ and the function form of $S_{\tau\lambda}(z)$.}
\begin{equation}
S_{\tau \lambda}(z) \equiv \left\{
      \begin{array}{ll}
          z-\tau \lambda &\ \ \ \ \ z> \tau \lambda \\
          0 &\ \ \ \ \ |z|\leq \tau \lambda \\ 
          z+\tau \lambda &\ \ \ \ \ z< -\tau \lambda
      \end{array}
      \right..
\end{equation}
It is now understood that the unit variance assumption in Eq. \eqref{Eq:BP} gives rise to the term $\norm{x-z}_2^2$ and thus corresponds to a convenient usage of the proximal mapping.

\subsection{ISTA and its limitation} 

The solution of the above fixed-point equation can be obtained via a natural iteration scheme:
\begin{equation}
        a_{l+1} = S_{\tau \lambda} (a_l-\tau K^T(Ka_l-g)),
    \label{Eq:Fixed-point-iteration}
\end{equation}
where $l=1,\ 2,\ \dots$ is the iteration step. Eq. \eqref{Eq:Fixed-point-iteration} is also called the iterative shrinkage-thresholding algorithm (ISTA) \cite{Daubechies2004,Figueiredo2003}. The optimizer ISTA is convex and usually guaranteed to find the global minimal of the BP problem, but for analytic continuation, it fails to converge to the physical fixed point because of the ill-poseness of the Fermi kernel.

To see this, we introduce the concept of mutual coherence $\mu$ to measure the ``fineness" or ill-poseness of the kernel matrix \cite{Donoho03,Gribonval03}. It is defined as the largest inner product between any two normalized columns of the matrix $K=[k_1|\dots|k_n] \in \mathbb{R}^{N_\tau \times N_\omega}$:
\begin{equation}
  \mu(K) \equiv \max_{i\neq j} \Big| \left\langle \frac{k_i}{\norm{k_i}_2},\frac{k_j}{\norm{k_j}_2}\right\rangle \Big|.
\end{equation}
For the BP problem \eqref{Eq:BP}, it has been proven \cite{Donoho03,Gribonval03} that the spectral function $a$ can be recovered exactly if it is sufficiently sparse, namely, $\norm{a}_0\leq \frac{1}{2}\left( 1+\frac{1}{\mu(K)} \right)$. Obviously, we have $0\leq \mu\leq 1$ for any real matrix. It achieves the lower bound 0 for an orthogonal matrix, but for the Fermi kernel matrix, it is straightforward to show that $\mu$ almost reaches the upper bound $1$. Thus, the corresponding BP problem is only guaranteed to recover the spectral function $a$ if it has just one non-zero entry. By contrast, a random kernel matrix whose columns are randomly sampled from a sphere can achieve a much smaller mutual coherence ($\mu \sim \sqrt{\frac{\log N_\omega}{N_\tau}}$) \footnote{This can be calculated from the inner product between two random points uniformly distributed on the unit sphere.}, so that its related BP problem can recover denser spectral functions. This raises a fundamental difficulty of the sparsity-based methods, if the spectral function is not sufficiently sparse and the Fermi kernel is not fine enough, namely, the mutual coherence is higher than needed. As a result, the ISTA optimizer will converge to unphysical solutions. To overcome this issue and enhance the power of the sparsity method, one strategy is to transform the BP problem to another one by multiplying a matrix on both sides of the inverse problem as proposed in Ref. \cite{Otsuki2017}. In this work, we explore a different strategy and deal with the ill-poseness of the Fermi kernal matrix by deep learning.

\subsection{Learned optimizers}
To overcome the issue of ill-poseness, we note that ISTA can also be viewed as recurrent neural networks (RNNs) with fixed weights determined by the Fermi kernel  matrix. A general RNN is a function modeling sequential data $\{x_l\}_{l=1}^L$ with parametrized function $x_{l+1}=f(x_l; \theta)$, where $l$ is the time step and $\theta$ is the collection of weights to be learned. The ISTA equation \eqref{Eq:Fixed-point-iteration} is nothing but a RNN with $x_l \rightarrow a_l$ and $f(x_l;\theta) \rightarrow S_{\tau \lambda}(a_l-\tau K^T(Ka_l-g))$, where the weights are fixed and given by the Fermi kernal matrix $K$. This immediately motivates us to design a neural network structure by unrolling the fixed-point iteration, namely, converting each iteration step into a single layer of the neural network and using the soft thresholding function $S$ as the activation function. We can then relax the fixed weights to be layer-dependent, utilize the simple forward problem of Eq. \eqref{Eq:Linear-AC} to generate data, train the weights for all layers, and learn an adaptive optimizer, in  hope that the learned weight matrices may be nicer.

The above line of thought leads to the following learnable iterative soft thresholding algorithm (LISTA) and its relaxation variation (RLISTA) with a $L$-layer neural network of fixed depth:
\begin{equation}
    \begin{split}
        a_{l+1} &= (1-\eta)a_l  + \eta S_{\tau \lambda} (W_t^l a_l + W_e^l g), \\
        a_{l+1} &= S_{\tau \lambda} (W_t^l a_l + W_e^l g),~~~l=1,2,\dots,L,
    \end{split}
\end{equation}
where $a_{l}$ and $a_{l+1}$ are the input and output for the $l$-th layer of the neural network respectively, $a_1$ and $g$ are inputs of the neural network and $0< \eta <1$ is the relaxation factor. The parameters $W_e^l$ and $W_t^l$ represent the layer-dependent weights to be learned on the $l$-th layer to replace the fixed weights $W_e=\tau K^T$ and $W_t=I-\tau K^TK$ in the original problem. During the training and inference processes, the neural networks are fed with the Green's function $g$ and a zero vector as $a_1$. 

\begin{figure}[t]
  \centering
  \includegraphics[width=0.45\textwidth]{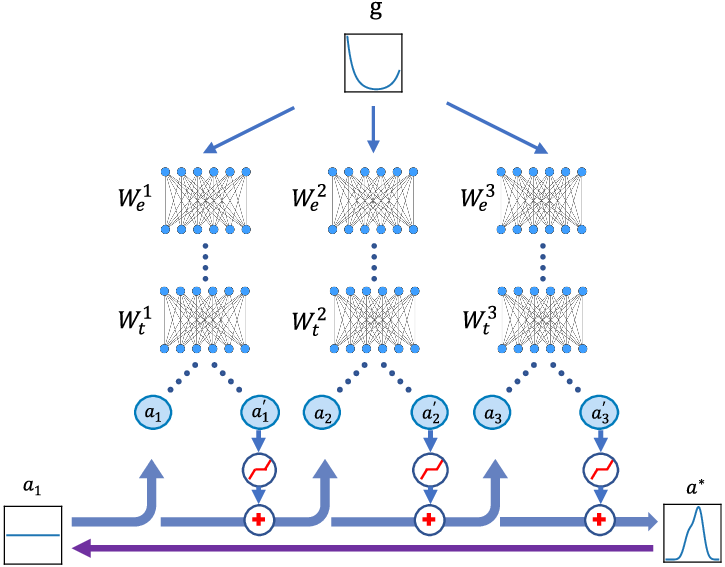}
  \caption{Architecture of the 3-layer RLISTA network with input of the Green's function $g$ and a zero vector $a_1$. The blue arrows illustrate the forward propagation of the  optimization process, and the purple arrow indicates the backward propagation which tunes all parameters to feed the data.}
  \label{fig:RLISTA}
\end{figure}

Unlike usual FCNs, our (R)LISTA network has no bias term and each LISTA layer has two matrices instead of one in common FCNs. Furthermore, as shown in Fig. \ref{fig:RLISTA}, RLISTA contains a residual connection \cite{He2016} term $(1-\eta)a_l$, so that the output of $(l+1)$-th layer is a linear combination of the output of $l$-th layer and a  LISTA layer. We call this neural network RLISTANet where the prefix $R$ refers to the relaxation or residual connection. Such constructive approach for solving inverse problems via neural network has made success in signal processing \cite{Gregor2010}.

\section{Results and discussions}
\subsection{Dataset generation}
To train the (R)LISTANet, we generate a dataset of $(g,a)$ with 100000 training samples and 10000 testing samples using the forward problem Eq. \eqref{Eq:Linear-AC}, where $a$ is obtained from the probability density function of Gaussian mixture distribution \cite{Fournier2020, Arsenault2017Projected}:
\begin{equation}
    A(\omega) = \frac{1}{N_R}\sum_{i=1}^{N_R} \exp\left[-\frac{(\omega-\mu_i)^2}{2\sigma_2^2}\right].
\end{equation}
Here, $N_R$ is the number of peaks valued in $\{1,\dots, 10\}$, $\mu_i\in [-1.5, 1.5]$ is the center of $i$-th peak, and $\sigma_i \in [0, 0.5]$ is the broadening. All three parameters $N_R$, $\mu_i$, $\sigma_i$ are random variables with uniform probability distribution. The frequency range is set to $\Omega_0=3$ so that the spectral function is only nonzero for $\omega \in [-\Omega_0,\Omega_0]$. In addition, we also add a quasiparticle peak centered near $\omega =0$, with $\mu_{\rm{center}} \in (-0.05,0.05)$ and $\sigma_{\rm{center}} \in (0.05,0.3)$. All generalized discretized spectral functions $a$ (as a vector) are normalized in the dataset with $N_\omega = 50$ and $N_\tau = 100$, which is a highly biased generation since the dimension of the Green's function (as a vector) is larger than that of the spectral function (as a vector). In this setting, the maximum entropy methods work better than the setting $N_\tau \le N_\omega$ for the purpose of comparison.

To simulate the effect of noise, the Green's functions are generated via 
\begin{equation}
    g = Ka + \sigma \odot \xi,
        \label{Eq:noise}
\end{equation}
where $\xi$ is the Gaussian white noise, $\sigma \in \{10^{-5}$, $10^{-4}$, $10^{-3}\}$ represents different noise levels, and $\odot$ denotes Hadamard product, i.e., element-wise product for noise at each $\tau_i$. Our neural networks are implemented using Tensorflow \cite{Abadi2016tensorflow} and optimized by Adam optimizer \cite{Kingma2014}. After training, they can produce the spectral function deterministically for each given sample of the Green's function. More training details are given in Appendix \ref{App:Training}.

\subsection{Performance and parameter efficiency} 
\begin{figure}[t]
    \includegraphics[width=0.45\textwidth]{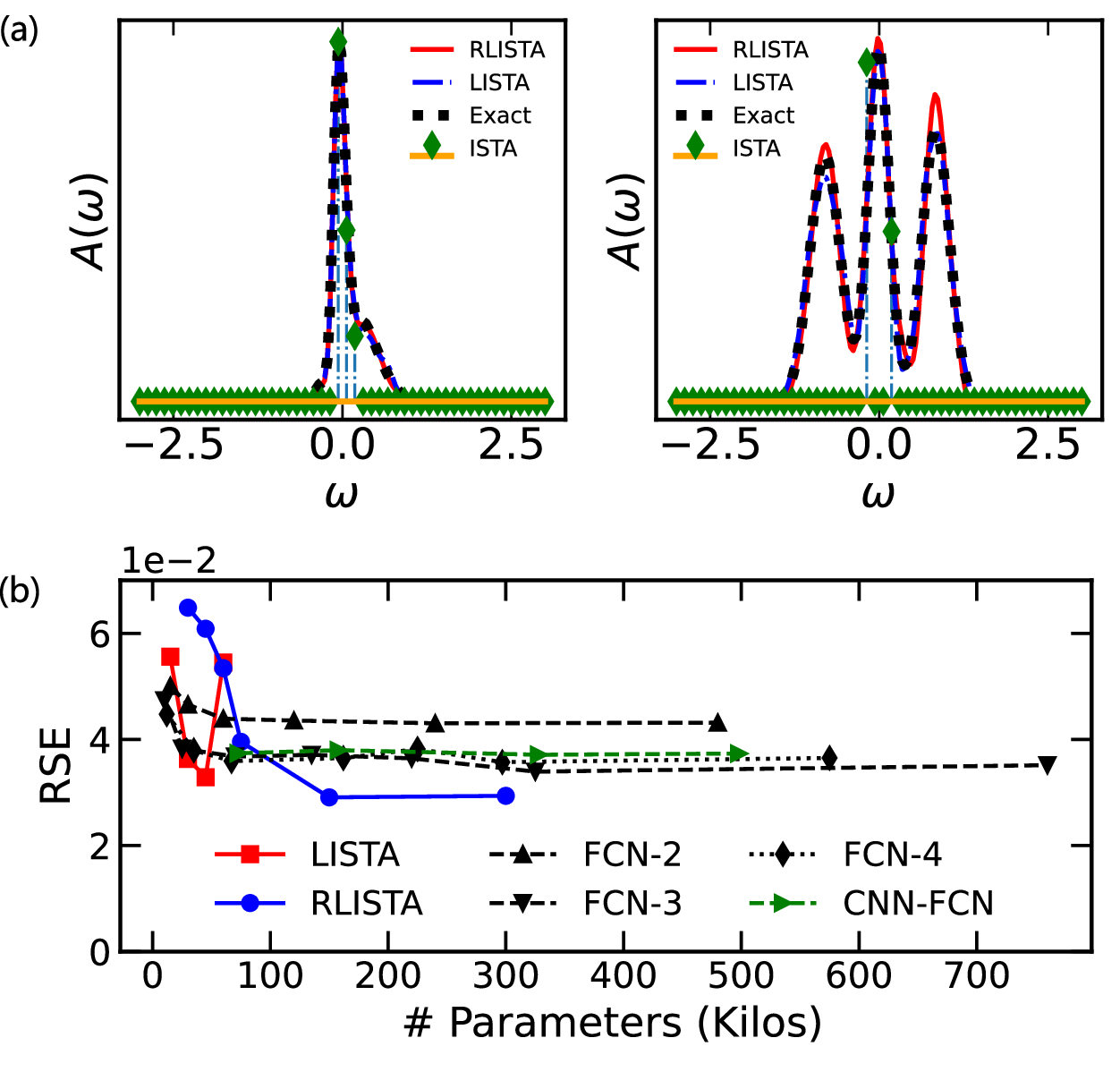}
    \caption{(a): Comparison of RLISTA, LISTA and ISTA for simulated data generated from the spectral function with only one sharp peak (left) and more peaks (right). The Green's function contains a noise level of $\sigma = 10^{-5}$. We have used arbitrary units for the vertical axis. (b) Comparison of parameter efficiency for different neural network architectures: LISTA, RLISTA, fully-connected network of two (FCN-2), three (FCN-3), or four (FCN-4) layers, and CNN-FCN networks. }
    \label{fig:Comparasion-Optimizer}
\end{figure}

Our learned optimizers, LISTA and RLISTA, have higher performance than vanilla ISTA under the small noise level, as compared in Fig. \ref{fig:Comparasion-Optimizer}(a). For simple spectra with only one sharp peak near the origin, all three optimizers can recover the solution well. But for complex spectra containing more broad peaks, ISTA can only give a single sharp peak, while both LISTA and RLISTA can produce the ground truth with high accuracy as measured by the root-square error (RSE):
\begin{equation}
  \text{\rm RSE}(\hat{a}) = \sqrt{\sum_{i=1}^{N_{\omega}} \left[\tilde{a}(\omega_{i}) - a^*(\omega_{i})\right]^2},
  \label{Eq:RSE}
\end{equation}
where $\tilde{a}$ is the ground truth spectral function in the test set and $a^*$ is the prediction of the optimizer. For the same task in Fig. \ref{fig:Comparasion-Optimizer}(a), ISTA, LISTA, and RLISTA give $\text{RSE}=0.155$, 0.016, 0.009 for single peak recovery and 0.24, 0.014, 0.02 for multi-peak recovery, respectively, showing an order of magnitude improvement in our learned optimizers.

Our neural networks may be viewed as a variation of the fully-connected network (FCN) but have higher parameter efficiency measured by RSE. This can be seen by comparison with the conventional FCN of one or two hidden layers and more advanced neural networks (4-layer FCN and 3-layer CNN-FCN) of varying width. For simplicity, the width of single FCN is set equal and given in Appendix \ref{App:Training}. As shown in Fig. \ref{fig:Comparasion-Optimizer}(b), (R)LISTA can achieve better accuracy than all others with several times more parameters. Of course, deeper CNN-FCN may have higher parameter efficiency if more convolutional layers are used. However, deep FCNs are known difficult to train and require more advanced techniques like normalization \cite{Ioffe2015,Ba2016,Wu2018}. 

\subsection{Residual connection in RLISTA}
One may notice in Fig. \ref{fig:Comparasion-Optimizer}(b) that the RSE of LISTA does not reduce monotonically with increasing number of parameters (layers). Hence, a deeper LISTANet may not necessarily outperform shallow ones, possibly due to the landscape of networks, namely, the loss function is highly non-convex and has many spurious local minima or large regions where the gradient directions do not point towards good minimizers \cite{Li2018}. By contrast, RLISTA contains residual connection (relaxation) \cite{He2016} and allows for the training of much deeper networks. Although shallow RLISTA cannot outperform LISTA, we can always train a deeper (up to 40 layers in our work) RLISTA that beats all other three, because residual connection is able to alleviate gradient vanishing \cite{He2016}, promote flat minima, and prevent the occurrence of high non-convexity when networks become deep \cite{Li2018}.

However, introducing residual connection may not completely remove the gradient vanishing problem. As seen in Fig. \ref{fig:Comparasion-Optimizer}(b), for the depth  larger than $20$, adding more layers can no longer reduce the RSE, reflecting a possible bottleneck of RLISTA. To understand this, we notice $a_{l+1} = (1-\eta) a_l + F_\theta(a_l,g)$, where all remaining terms are denoted as $F_\theta(a_l)$ for simplicity. Hence, for an $L$-layer RLISTANet ($l<L$), we have
\begin{equation}
a_{L} = \left[ \prod_{i=l}^{L-1}(1-\eta) \right]  a_l + \sum_{i=l}^{L-1}F_\theta(a_i,g).
\end{equation}
During backpropagation, the gradient of loss function with respect to the $l$-th layer can then be evaluated from the chain rule: 
\begin{equation}
  \frac{\partial L}{\partial a_l} = \frac{\partial L}{\partial a_L} \left[ (1-\eta)^{L-l} + \frac{\partial }{\partial a_l}\sum_{i=l}^{L-1}F_\theta(a_i) \right].
  \label{Eq:Gradient}
\end{equation}
The above equation illustrates a decomposition of gradient: the first term, $\frac{\partial L}{\partial a_L}(1-\eta)^{L-l}$, propagates information of the target spectral function directly without considering any intermediate layers, and the second term, $\frac{\partial L}{\partial a_L} \frac{\partial }{\partial a_l}\sum_{i=l}^{L-1}F_\theta(a_i)$, propagates the same information through intermediate layers. As long as $L-l$ is not too large, the first term is finite and would not always be canceled by the second term in a batch. Thus, the gradient of these layers will not vanish and the network can be effectively trained. However, for an ultra deep neural network ($L \gg 1$), the first term is exponentially small, $\sim e^{(L-l)\ln(1-\eta)}$, for shallow layers $(l\ll L)$, while the second term is empirically hard to control and might cause gradient vanishing \cite{He2016b}, so that these layers cannot be trained effectively and only a limited number of layers can be utilized for an ultra deep RLISTANet. This defines an "effective depth" of RLISTANet determined by the relaxation factor $\eta$. A thorough investigation of the relation between residual connection and the bottleneck of RLISTANet is beyond the scope of this work.

The above analysis also indicates that the residual connection or relaxation can improve the trainability of our learned optimizer. However, most of previous studies on the relaxation methods have focused on unlearned fixed-point problems to accelerate convergence and improve stability. It is therefore interesting to speculate if results on the acceleration and convergence conditions may be extended to the learned networks and help design better architecture and optimization techniques with improved performance and higher accuracy.

\begin{figure}[t]
    \includegraphics[width=0.45\textwidth]{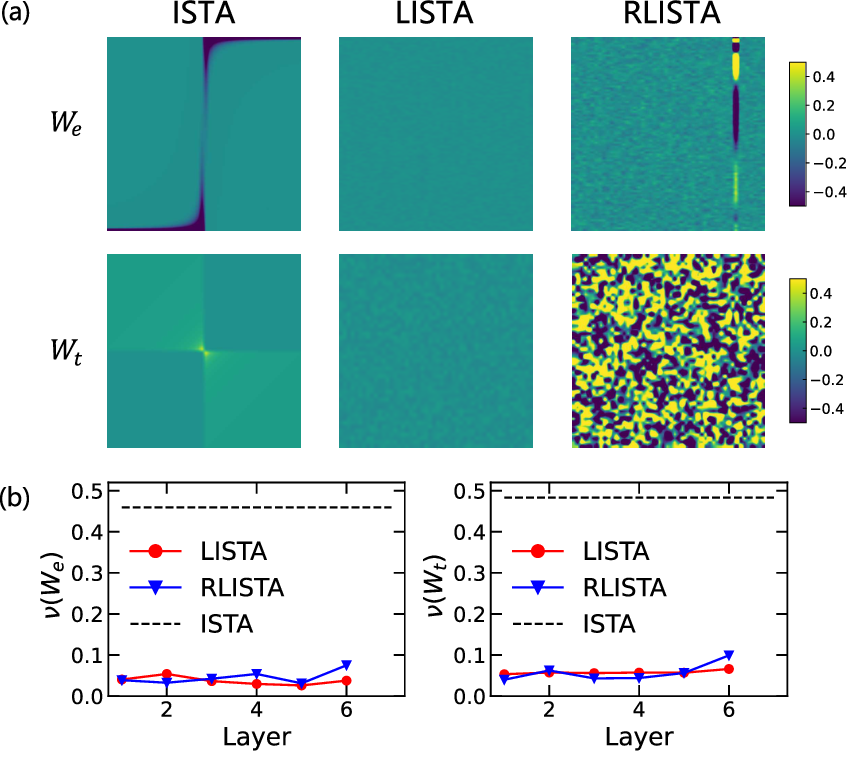}
    \caption{(a) Visualization of the first layer matrices $W_t$ and $W_e$ in ISTA, LISTA and RLISTA optimizers. Each pixel represents an entry of the matrix. The magnitude of the entry is shown by the color bar. The entries are normalized by their Frobenius norm $\|W_e\|_F^2$ and $\|W_t\|_F^2$. Before normalization, an identity matrix is subtracted from the matrix $W_t$, while the learned $W_e$ and all matrices related to ISTA are multiplied by $-1$. For $W_e$, we have  a mean value of 3.5$\times10^{-5}$ and a variance of 1.2$\times10^{-4}$ for LISTA, and a mean value of -3.3$\times 10^{-3}$ and a much larger variance of 5.3$\times 10^{-2}$ for RLISTA. For $W_t$, the mean and variance are -7.8$\times10^{-4}$ and 5.1$\times10^{-4}$ for LISTA, and -3.6$\times10^{-3}$ and 9.8$\times10^{-1}$ for RLISTA. (b) Comparison of the average coherence of their learned matrices $\nu(W_e),\nu(W_t)$ for all layers, showing the benefits of learning in LISTA and RLISTA.}
    \label{fig:Weight matrix}
\end{figure}

\subsection{The weight matrices} 
What is the reason behind the substantial improvement in learned optimizers? Noticing that ISTA may be viewed as an infinite-layer forward neural network with fixed weight because it typically needs infinite iterations to approach the exact fixed point, the improvement of (R)LISTA must originate from the adaptation to the data as reflected in the learned weight matrices. After training, like all other machine learning methods, our learned optimizer should be able to recover a spectral function well if similar data have been included in the training set. The quality of recovery can be improved by building a larger training set and using a deeper and wider neural network \cite{Yoon2018}.

Figure \ref{fig:Weight matrix}(a) visualizes an example of the normalized learned weight matrices $W_t$ and $W_e$ on first layer for different optimizers. While the matrices in ISTA are determined directly by the Fermi kernel, the learned matrices in LISTA and RLISTA are heavily influenced by their different iteration schemes. All entries of $W_e$ and $W_t$ in LISTA are distributed uniformly and are of the same magnitude, while those in RLISTA differ heavily and contain some entries of relatively larger values, as manifested by the much larger variances than those for LISTA. Nevertheless, both types of learned matrices can empirically solve the inverse problem and perform nicer than their vanilla cousin ISTA.

To quantify the “niceness” of the learned matrices, we use the average coherence \cite{Bajwa2010,Mixon2011} for any matrix $B=[b_1|\dots|b_n] \in \mathbb{R}^{m\times n}$:
\begin{equation}
    \nu(B) = \frac{1}{n-1}\max_{i\in\{1,\dots,n\}} \Big| \sum_{j\neq i, j=1}^n \left\langle \frac{b_i}{\norm{b_i}_2},\frac{b_j}{\norm{b_j}_2} \right\rangle \Big|,
\end{equation}
which gives the largest average inner product of two columns and measures the spread of column vectors of a matrix within a unit ball. The average coherence takes value in $[0,1]$, becomes zero for an orthogonal matrix, and reaches its upper bound for a matrix with repeated columns. Obviously, a lower average coherence implies a ``nicer" or less singular matrix. As shown in Fig. \ref{fig:Weight matrix}, the average coherence of learned matrices is around the lower bound zero, while the predefined matrices in ISTA have a large average coherence close to 0.5. The gap between them illustrates the benefits of learning.

The diversity of learned matrices in different optimizers might be understood as the diversity of learned problems. Each layer of LISTA and RLISTA networks works approximately as an independent iteration step for solving a given BP problem of Eq. \eqref{Eq:BP} with their individual learned kernel matrix. But a single iteration can not solve the problem with high accuracy. With many learned layers, the neural networks learn a sequence of optimization problems and solve each of them approximately, in the sense that only one iteration is performed  on each layer while an exact solution of the BP problem requires many iterations to achieve convergence.

From the perspective of optimization, our method may be understood as a generalization of the classical homotopy method \cite{Donoho2008} which solves the BP problem by a sequence of problems with varying regularization parameter $\lambda$. Here the neural network solves the problem by a series of problems with varying $K$ and $\lambda$ in the region of physical Green's functions, which is much smaller than the whole space $\mathbb{R}^{N_\tau}$. Thus, a trade-off comes from the fixed depth: how precisely do we need to solve a single question (precision) and how many optimization problems do we need to learn (diversity)? Careful studies of the precision-diversity trade-off may reveal a closer relationship between optimization and neural network, which we leave for future work.

\subsection{Robustness and stability of recovery}
The presented results in previous figures are obtained for a small noise $\sigma=10^{-5}$. However, our networks are robust and stable under different noise levels $\sigma \in \{10^{-5}$, $10^{-4}$, $10^{-3}\}$. To  see this, we train the neural networks in a \textit{noiseless} dataset and test the trained networks for each noise level with $200$ \textit{noisy} Green's functions generated randomly using Eq. \eqref{Eq:noise} for a fixed spectral function $a$. The robustness and stability against noise can be quantified by the standard variance (Std) of the output spectral function and the RSE between average result predicted by neural network and the exact spectral function. While the former (Std) measures the noise sensitivity (robustness) of the neural network, the latter (RSE) qualifies the stability of the output.

\begin{figure}[t]
  \centering
  \includegraphics[width=0.45\textwidth]{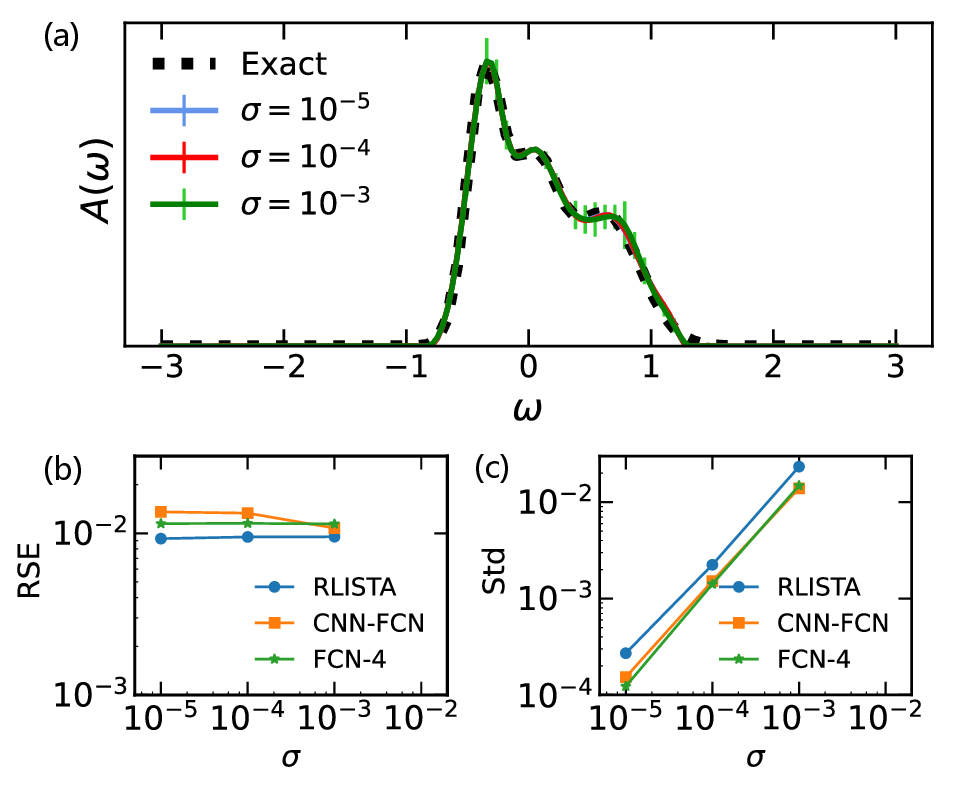}
  \caption{Robustness and stability tests for RLISTA, CNN-FCN, FCN-4 networks. (a) An example of spectral function recovery (in arbitrary units) obtained by RLISTA under different noise levels $\sigma \in \{10^{-5}$, $10^{-4}$, $10^{-3}\}$. (b) Stability measured by the root-square-error (RSE) for different neural networks as a function of the noise level. (c) Robustness measured by the standard variance (Std) of the output for different neural networks as a function of the noise level.}
  \label{fig:robustness}
\end{figure}

As an example, we test a 10-layer RLISTA network with 75000 parameters and the results are compared in Fig. \ref{fig:robustness} with two other baseline networks. The first one is CNN-FCN \cite{Yoon2018} with 50 neurons in the first layer and 32 channels in the second convolutional layer to ensure similar number of parameters as RLISTA. The second one is a four-layer network of the width 256, which has 169000 parameters and was reported to achieve good performance \cite{Fournier2020}. As shown in Fig. \ref{fig:robustness}(a), RLISTA is able to recover the spectral function for all three noise levels. The quality of recovery as measured by RSE shown in Fig. \ref{fig:robustness}(b) is even slightly better than that of CNN-FCN and FCN-4. On the other hand, Fig. \ref{fig:robustness}(c) shows a slightly larger Std of RLISTA than CNN-FCN and FCN-4, indicating that RLISTA might be slightly more sensitive to noise. However, if we have enough noisy samples of the Green's function and take average over their output spectral functions, RLISTA is still able to work no worse than CNN-FCN or FCN-4 as manifested by its smaller RSE.

\subsection{Neural default model}
The neural networks can give an answer with low time cost after training, but they are often criticized as being a “black-box”, where precise interpretations of the weights remain unclear. Better neural network architectures and optimization techniques can no doubt reduce the error, but their design suffers from high cost trial and error. On the other hand, a classical maximum entropy problem requires a default model $d(\omega)$ to incorporate certain prior knowledge about the desired spectral function. A better default model allows for easy hyperparameter tuning and gives better accuracy. We propose that combining the two may provide a novel way to improve the performance. The output of our neural network is also a probability distribution and may be considered as a ``neural default model" required in the maximum entropy. This can fix the inexactness of the spectral function predicted  by the neural networks and benefit from both the high speed of neural networks and the well-developed algorithms of the maximum entropy.

\begin{figure}[t]
  \centering
  \includegraphics[width=0.45\textwidth]{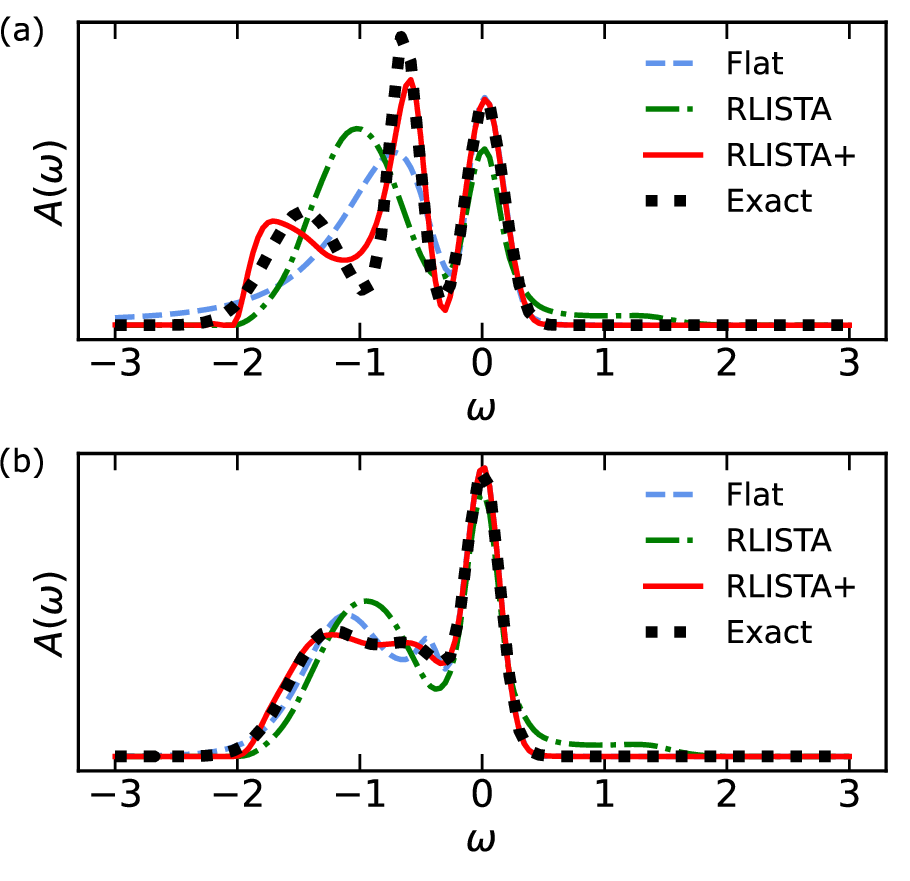}
  \caption{Two examples of the spectral function recovery (in arbitrary units) under small noise $(\sigma=10^{-5})$ using the maximum entropy method with a flat default model (Flat, dashed line) and a neural default model (RLISTA+, red solid line). The dotted-line is the ground truth spectral function. The RLISTA results (green dot-dash line) are generated by a sub-optimal 6-layer network. For both examples, the error bars are smaller than the line width.}
  \label{fig:Neural+}
\end{figure}

To get an impression on the advantage of such combination, we first obtain an inexact result from a sub-optimal 6-layer RLISTA under small noise ($\sigma=10^{-5}$). Figure \ref{fig:Neural+} shows two examples where both RLISTA (green dot-dash line) and the maximum entropy with flat or uniform prior (blue dashed line) cannot capture all the details of the spectral function. As shown in Figure \ref{fig:Neural+}(a), while RLISTA tends to mix two peaks and give an average peak, the maximum entropy method ignores the peak at high frequency. By contrast, using the neural default model, the maximum entropy solution RLISTA+ (red solid line) can capture well the high frequency peak as in Fig. \ref{fig:Neural+}(a) or even all the details of the exact spectral function as in Fig. \ref{fig:Neural+}(b). This neural maximum entropy method (RLISTA+) makes a promising improvement over the conventional one for analytic continuation. 

\section{Conclusion}
Motivated by sparsity-based methods, we have proposed a highly efficient neural network scheme for analytic continuation in quantum many-body problems. Our learned optimizers show low time costs and may be easily extended to other high-dimensional inverse problems via large-scale pretraining where traditional maximum entropy methods demand tremendous computation time (see Appendix \ref{App:Benchmarks} for more details). The output of our method may also be used as a neural default model to improve the performance of maximum entropy methods and make use of their both advantages. We also find that constructing neural networks from fix-point iteration can achieve better parameter efficiency than heuristic fully-connected networks. By viewing neural network as learnable fix-point iteration, we see that different fix-point iteration schemes are not equivalent if their parameters can be learned, despite that they all converge to the same solution for a given problem. The learned iteration paths are more regular than their unlearned counterpart. This, combined with the powerful theory about calculus of variations, may help invent novel algorithms and lead to a better understanding of the training dynamics and regularization of neural networks. 
 
This work was supported by the National Natural Science Foundation of China (NSFC Grants No. 11974397, No. 12174429), the National Key R\&D Program of MOST of China (Grant No. 2017YFA0303103), the Strategic Priority Research Program of the Chinese Academy of Sciences (Grant No. XDB33010100), and the Youth Innovation Promotion Association of CAS.

\appendix

  \section{Neural network architecture and training details}
  \label{App:Training}
  We use the RSE loss function to train all the neural networks $\mathcal{L} = \sqrt{\|a_\theta - \tilde{a}\|_2^2}$, where $a_\theta$ is the output of the neural network and $\tilde{a}$ denote the ground truth. Further, we implement the neural network in Tensorflow \cite{Abadi2016tensorflow} and optimize the weights by Adam optimizer with exponential decay learning rate. All results are obtained in a desktop PC with AMD Threadripper 2950x CPU and NVIDIA 2080 TI GPU. Table \ref{Tab:Training Hyperparameter} shows the general training hyperparameters used in this work. The decay step is about $10$ epoches in our setting.
  \begin{table}[ht]
    \caption{Training hyperparameters.}
    \label{Tab:Training Hyperparameter}       
    
    
    \begin{tabular}{ll}
    
    \hline\noalign{\smallskip}
    
    Hyperparameter & Value\\
    
    \noalign{\smallskip}\hline\noalign{\smallskip}
    
    Epoch & $150$ \\

    Batch Size & $256$\\
    
    Learning Rate &$0.01$ \\

    Decay Step & $3906$ \\

    Decay rate  & $0.9$ \\    
    \noalign{\smallskip}\hline
    \end{tabular}
\end{table}

\begin{figure}[h]
  \centering
  \includegraphics[width=0.45\textwidth]{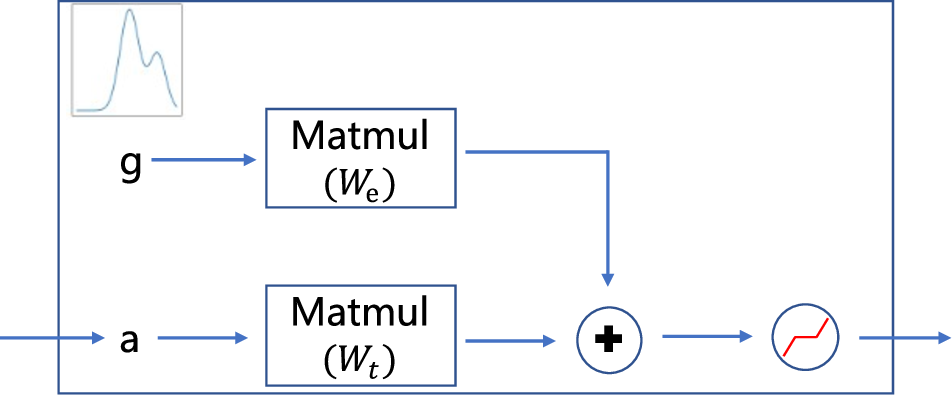}
  \caption{The semantic diagram for a LISTA layer.}
  \label{fig:LISTA}
\end{figure}

  A layer of LISTA contains two inputs, two matrices, and one element-wise activation function. The architecture is shown in Table \ref{Tab:LISTA} and Fig. \ref{fig:LISTA}. The depth is chosen in the set $\{4,5,6,7\}$ for the benchmark of parameter efficiency. $\ell^2$ regularization is used except in the last layer.
  \begin{table}[ht]
    \caption{Architecture of the $l$-th layer in LISTANet.}
    \label{Tab:LISTA}       
    
    
    \begin{tabular}{lll}
    
    \hline\noalign{\smallskip}
    
    LISTA \\
    
    \noalign{\smallskip}\hline\noalign{\smallskip}
    
    Input Green's function $g\in \mathbb{R}^{N_\tau}$ & \\
    
    Last layer's output $a_{l-1} \in \mathbb{R}^{N_\omega}$ \\

    Matrix $W_e \in \mathbb{R}^{N_\omega \times N_\tau}$ \\

    Matrix $W_t \in \mathbb{R}^{N_\tau \times N_\tau}$  \\

    Soft-thresholding function with learnable parameter$^{a} S_\lambda(.)$ \\

    $\ell^2$ regularization factor $0.01$ \\ 
    
    \noalign{\smallskip}\hline
    
    \end{tabular}
    \tablenote{$\lambda$ is shared for different neurons in the layer.} 
\end{table}

\begin{figure}[ht]
  \centering
  \includegraphics[width=0.45\textwidth]{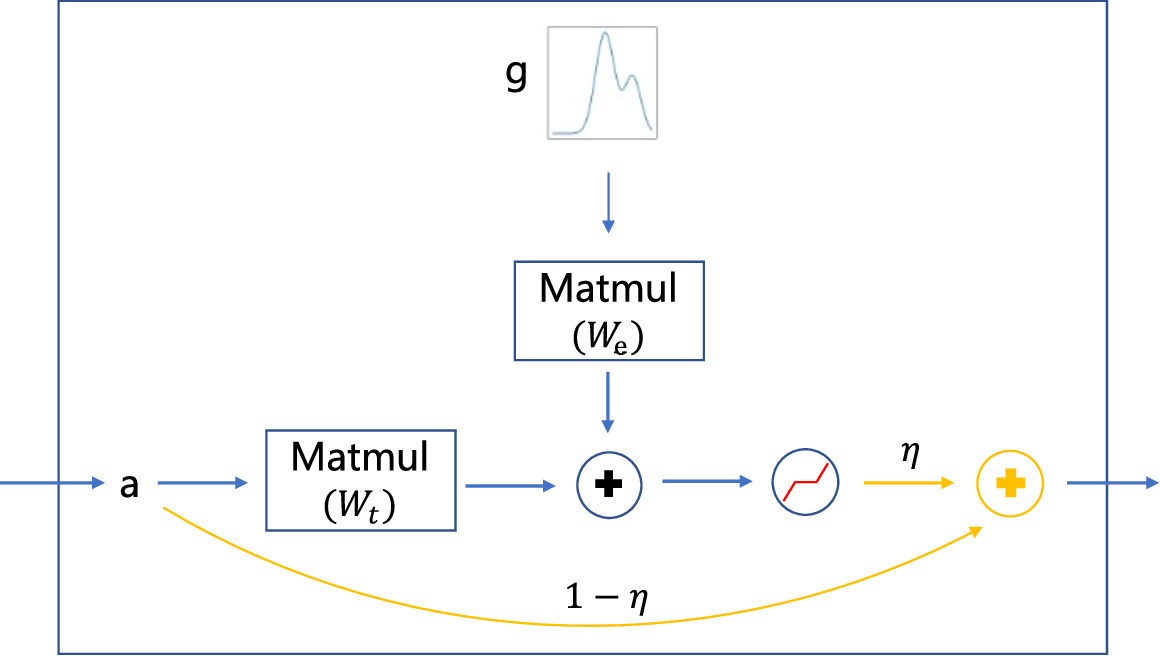}
  \caption{The semantic diagram for a RLISTA layer.}
  \label{fig:ARLISTA}
\end{figure}

The architecture of RLISTA is similar to that of LISTA and shown in Table \ref{Tab:RLISTA} and Fig. \ref{fig:ARLISTA} with additional residual connection. We choose the relaxation factor $\eta  = \frac{1}{2}$ and perform LeCun initialization\cite{lecun98b} in each layer, namely, all entries satisfy $w_{ij} \overset{i.i.d.}{\sim} \mathcal{N}(0, \frac{1}{N_\tau})$ for the matrix $W_e$ and $w_{ij} \overset{i.i.d.}{\sim} \mathcal{N}(0, \frac{1}{N_\omega})$ for the matrix $W_t$. For the benchmark of parameter efficiency, the depth is chosen in the set $\{4,6,8,10,20,40\}$. Additionally, $\ell^2$ regularization is used except in the last layer.

  \begin{table}[h]
    \caption{Architecture of the $l$-th layer in RLISTANet.}
    \label{Tab:RLISTA}       
    
    
    \begin{tabular}{lll}
    
    \hline\noalign{\smallskip}
    
    RLISTA \\
    
    \noalign{\smallskip}\hline\noalign{\smallskip}
    
    Input Green's function $g\in \mathbb{R}^{N_\tau}$ & \\
    
    Last layer's output $a_{l-1} \in \mathbb{R}^{N_\omega}$ \\

    Matrix $W_e \in \mathbb{R}^{N_\omega \times N_\tau}$ \\

    Matrix $W_t \in \mathbb{R}^{N_\tau \times N_\tau}$  \\

    Soft-thresholding function with a learnable parameter$^{a} S_\lambda(.)$ \\

    Relaxation factor $\eta = \frac{1}{2}$ \\
    $\ell^2$ regularization factor $0.01$ \\ 
    \noalign{\smallskip}\hline
    
    \end{tabular}
    \tablenote{$\lambda$ is shared for different neurons in the layer.} 
\end{table}

FCNs are trained in a supervised learning manner to fit the exact spectral function given a corresponding Green's function. An equal width $w$ is set for each FCN and chosen from the set $\{100, 200, 400, 800, 1600, 3200\}$ for FCN-2, $\{50, 100, 200, 300, 400, 500, 600\}$ for FCN-3, and $\{50, 100, 150, 250, 300, 350, 500\}$ for FCN-4. The detailed architecture is listed in Table \ref{Tab:FCN-2}.

\begin{table}[h]
  \caption{Architecture of each layer for FCNs.}
  \label{Tab:FCN-2}       
  
  
  \begin{tabular}{lllll}
  
  \hline\noalign{\smallskip}
  
  type & input size & output size  & activation function \\
  
  \noalign{\smallskip}\hline\noalign{\smallskip}
  
  FC (first layer) & $N_\tau$ & $w$  & ReLU\\
  FC & $w$ & $w$   & ReLU\\
  FC (last layer) & $w$ & $N_\omega$   & softmax\\
  \noalign{\smallskip}\hline
  
  \end{tabular}
  
\end{table}

CNN-FCN is also trained in a supervised learning manner. The network receives a Green's function and tries to fit the exact spectral function. The CNN-FCN network has a fully-connected layer, followed by a 32 channel convolutional layer and a fully-connected last layer with softmax activation for normalization. For parameter efficient test, the width is chosen in the set $\{50, 100, 200, 300\}$.  The detailed architecture is listed in Table \ref{Tab:CNN_FCN}.

\begin{table}[h]
  \caption{Architecture of CNN-FCN.}
  \label{Tab:CNN_FCN}       
  
  
  \begin{tabular}{lllllll}
  
  \hline\noalign{\smallskip}
  
  type & kernel & stride & input size & output size  & \begin{tabular}{l}
    activation \\ function
  \end{tabular}\\
  
  \noalign{\smallskip}\hline\noalign{\smallskip}
  
  FC & & & $N_\tau$ & $w$   & ReLU\\
  conv1d & $8\times 32$ &1 & $w$ & $32\lceil \frac{w-8+1}{1} \rceil$   & ReLU\\
  FC & & & $32\lceil \frac{w-8+1}{1} \rceil$ & $N_\omega$   & softmax\\
  \noalign{\smallskip}\hline
  
  \end{tabular}
  
\end{table}

\section{Benchmark for time cost}
\label{App:Benchmarks}
To test the time cost of our neural networks, we show in Table \ref{Tab:Time-cost} some results of high-dimensional extensions recovering 10000 spectral functions for $N_\omega = 100$, 200, 400 with a fixed $N_\tau/N_\omega = 2$. Four architectures are compared: 6-layer RLISTA, 20-layer RLISTA, 4-layer FCN of width 256, and CNN-FCN whose first layer has a width $w=500$ and second layer is a one-dimensional convolutional layer with the kernel $8 \times 32$ (stride=1). We use ReLU function for CNN-FCN in the first two layers and softmax function in the last layer for normalization. We see that the time cost is not sensitive to dimensionality for $N_\tau \leq 800$ and $N_\omega \leq 400$.

\begin{table}[h]
  \caption{Time cost (in seconds) of different architectures.}
  \label{Tab:Time-cost}       
  
  \setlength\tabcolsep{7pt}
  \begin{tabular}{llllllll}
  
  \hline\noalign{\smallskip}
  
  Network  & $N_\omega = 100$ &$N_\omega = 200$ &$N_\omega = 400$ \\
  
  \noalign{\smallskip}\hline\noalign{\smallskip}
   6-RLISTA & $0.30\pm 0.02$  & $0.30\pm 0.01$ &$0.35\pm 0.03$ \\
   20-RLISTA & $0.44\pm 0.01$  &$0.46\pm 0.02$ &$0.53\pm 0.02$ \\
   CNN-FCN & $1.37\pm 0.04$  &$1.34 \pm 0.02$ &$1.38\pm 0.02$ \\
   FCN-4 &$0.24\pm 0.01$ &$0.26 \pm 0.01$ &$0.29\pm 0.01$ \\
   MaxEnt &\multicolumn{3}{c}{$\sim\,$3 hours for $N_\omega = 50$}\\

  \noalign{\smallskip}\hline
  
  \end{tabular}
  
\end{table}

\end{document}